\documentclass[10pt,twocolumn,letterpaper]{article}

\usepackage[pagenumbers]{cvpr} 

\usepackage{multirow}
\usepackage{pifont}
\newcommand{\cmark}{\ding{51}}
\newcommand{\xmark}{\ding{55}}

\usepackage[dvipsnames]{xcolor}

\definecolor{cvprblue}{rgb}{0.21,0.49,0.74}
\usepackage[pagebackref,breaklinks,colorlinks,citecolor=cvprblue]{hyperref}

\def\paperID{16307} 
\def\confName{CVPR}
\def\confYear{2024}

\title{Neural Text to Articulate Talk: Deep Text to Audiovisual Speech Synthesis achieving both Auditory and Photo-realism}

\author{\normalsize{Georgios Milis\textsuperscript{1} \quad Panagiotis P. Filntisis\textsuperscript{1,2} \quad Anastasios Roussos\textsuperscript{3} \quad Petros Maragos\textsuperscript{1,2}}  \\
\vspace{-0.4cm}
\\
\footnotesize{\textsuperscript{1}School of Electrical \& Computer Engineering, National Technical University of Athens, Greece}\vspace{-0.1cm}\\
\footnotesize{\textsuperscript{2}Institute of Robotics, Athena Research Center, 15125 Maroussi, Greece} \vspace{-0.1cm}\\
\footnotesize{\textsuperscript{3}Institute of Computer Science (ICS), Foundation for Research \& Technology - Hellas (FORTH), Greece}\vspace{-0.1cm}
\\
}

\begin{document}

\twocolumn[{%
\renewcommand\twocolumn[1][]{#1}%
\maketitle
\begin{center}
    \centering
    \captionsetup{type=figure}%
    \rotatebox{90}{~~[\textit{manuscripts}]}
    \rotatebox{90}{}
    \includegraphics[width=0.75\textwidth]{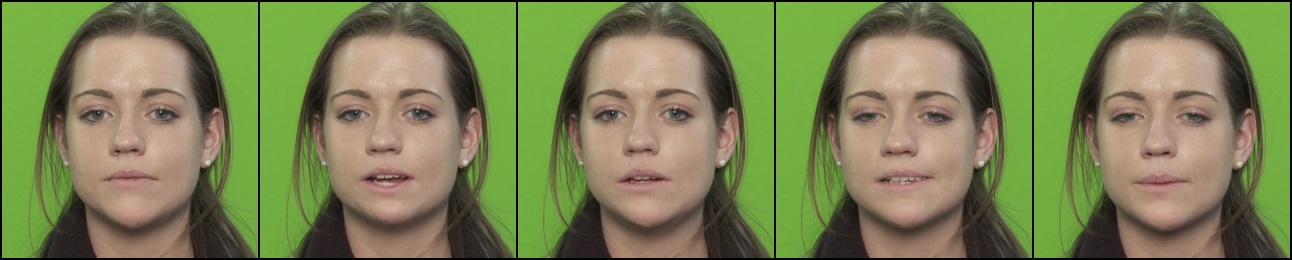}\\
    \rotatebox{90}{~~~~~[\textit{campfire}]}
    \rotatebox{90}{}
    \includegraphics[width=0.75\textwidth]{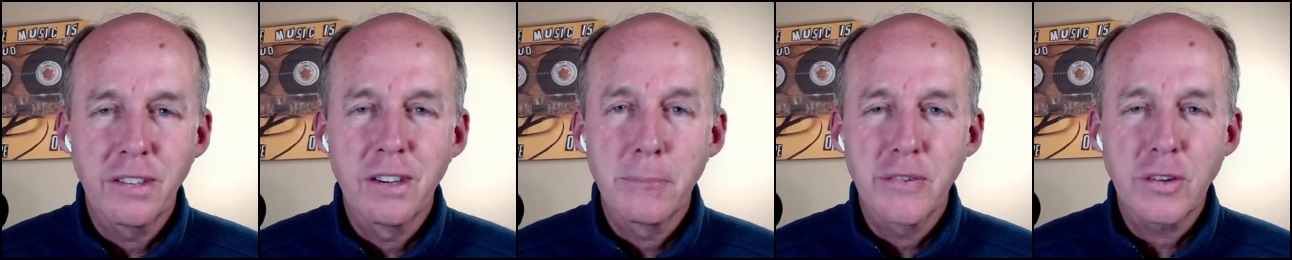}\\
    \rotatebox{90}{~~~~~[\textit{removal}]}
    \rotatebox{90}{}
    \includegraphics[width=0.75\textwidth]{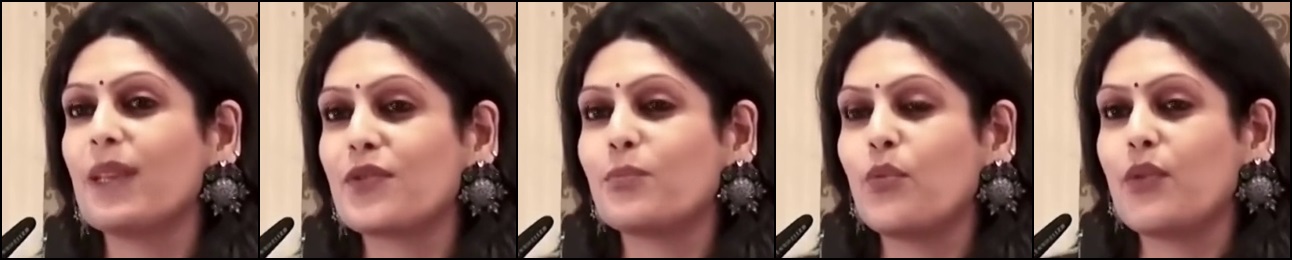}\\
    \rotatebox{90}{[\textit{fortune smiles}]}
    \rotatebox{90}{}
    \includegraphics[width=0.75\textwidth]{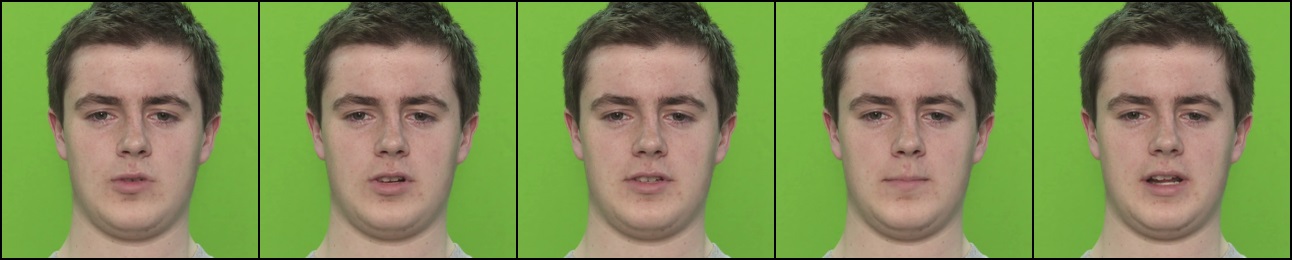}\\
    \captionof{figure}{Our method can produce photo-realistic videos of talking heads, with lifelike lip articulation. Realistic results can be achieved for talking heads captured either in lab conditions or in-the-wild. On the left we include the phrase uttered in each set of frames. Please zoom in to observe the subject's lip formations and refer to the project website \cite{project_page} more results.}
    \label{fig:samples}
\end{center}%
}]

\begin{abstract}
Recent advances in deep learning for sequential data have given rise to fast and powerful models that produce realistic videos of talking humans. The state of the art in talking face generation focuses mainly on lip-syncing, being conditioned on audio clips. 
However, having the ability to synthesize talking humans from 
text transcriptions rather than audio is particularly beneficial for many applications and is expected to receive more and more  attention, following the recent breakthroughs in large language models. 

For that, most methods implement a  cascaded 2-stage architecture of
a text-to-speech module followed by an audio-driven talking face generator, but this ignores the highly complex interplay between audio and visual streams that occurs during speaking. 
In this paper, we propose the first, to the best of our knowledge,
text-driven audiovisual speech synthesizer that uses Transformers
and does not follow a cascaded approach. 
Our method, which we call NEUral Text to ARticulate Talk (NEUTART), is 
a talking face generator that uses a joint audiovisual feature space, as well as speech-informed 3D facial reconstructions and a lip-reading loss for visual supervision. The proposed model produces photorealistic talking face videos with human-like articulation and well-synced audiovisual streams. Our experiments on audiovisual datasets as well as in-the-wild videos reveal state-of-the-art generation quality both in terms of objective metrics and human evaluation.
\end{abstract}
    
\section{Introduction}
\label{sec:intro}

Speech synthesis has been one of the pronounced successes of generative AI. State-of-the-art text-to-speech (TTS) systems' output is almost indistinguishable from real human speech \cite{tan2021survey}. While the sound of speech is its most important aspect, the visual component is equally essential for conveying meaning. Talking face generation has numerous applications, including digital avatars, virtual assistants, accessibility tools, teleconferencing, video games, movie dubbing, and human-machine interfaces \cite{toshpulatov2023talking}. 

Such generators can either be audio-driven or text-driven, with audio-driven models being more common. Text-driven audiovisual speech synthesis is underrepresented in research, which is surprising considering the prominence of models that use natural language input. A text-driven model is suitable for extending existing conversational language models that have recently exploded in popularity, by providing an animated avatar that utters the model's output. It can also contribute to building simpler animation pipelines for digital media professionals.

A common approach for text-driven models is to use a cascaded architecture of a TTS module and an audio-driven talking face generator \cite{kumar2017obamanet, wang2022anyonenet, song2022talking, zhang2022text2video, ye2023ada}. However, these 2-stage architectures have to drive the talking face generator with features from an audio feature extractor, which encodes the TTS system's synthesized speech into another speech-related representation. We claim that this is redundant, since the TTS system already uses an intermediate speech representation coming from the text. 

Other text-driven models focus only on generating visual speech from text \cite{li2021write, liu2022parallel}, leaving the synthesis of audio speech out of their scope, with obvious problems in terms of consistency and synchronization between the audio and visual modality. 
Finally, very few recent methods perform audiovisual speech synthesis from text \cite{yu2019durian, hussen2021audiovisual, mitsui2023uniflg}, but none of these methods has the ability to synthesize photorealistic videos. In addition, these methods either adopt autoregressive architectures that have limitations in terms of sampling speed and dealing with longer samples, or adopt an over-simplistic representation of facial movements based on sparse 2D landmarks. 

Our NEUTART method is the first, to the best of our knowledge, text-driven, photo-realistic audiovisual speech synthesizer
that is genuinely bimodal. 
In more detail, NEUTART uses a joint audiovisual representation, avoiding the need for a cascaded 2-stage approach. Leveraging a powerful Transformers-based architecture, we use 
supervision from both audio and visual speech during training, forcing the intermediate features to effectively encode the inherent bimodality of speech. 
We also adopt a dense 3D representation of the speech-related facial motions based on 3D Morphable Models (3DMMs) and use it to condition a neural renderer based on Generative Adversarial Networks (GANs), producing photorealistic facial videos. 

As a result, our model learns to synthesize realistic lip movements and combine them with natural head pose dynamics in a particularly photorealistic manner, see Fig.~\ref{fig:samples}. It also synthesizes speech audio that is especially consistent with the video, leading to audiovisual results of unprecedented realism and plausibility.   
This is reflected in the quantitative and qualitative results as well as the user study that we present in the experiments, providing evidence about the promising results and the  advantages of NEUTART over the previous methods. 
In summary, our main contributions are the following:
\begin{itemize}
\item 
We introduce the first, to the best of our knowledge, text-driven, photo-realistic audiovisual speech synthesizer that is genuinely bimodal and avoids the cascaded 2-stage approaches for audio and video synthesis adopted by previous methods
\item 
We propose a novel joint modeling of acoustic and 3D visual elements in a learned feature space, which captures the complex interplay between audio and visual streams. This increases the perceived realism and plausibility of the final synthetic result.
\item 
We adopt a detailed 3D representation for the synthesis of visual speech and combine it with state-of-the-art photo-realistic video synthesis based on conditional GANs. This 
allows us to blend the synthesized facial motions that match the input text with complex, challenging scenes in a completely photo-realistic manner, paving the way for a multitude of extended capabilities for AI-based video synthesis. 
\item 
We conduct qualitative and quantitative experiments, user  and ablation studies to evaluate our method and compare it with recent state-of-the-art methods. The experiments demonstrate the effectiveness and advantages of our method, which achieves particularly promising results in challenging scenarios. 
\item 
We make the source code of our method publicly available at the project's website \cite{project_page}.
\end{itemize}

\section{Related work}
\label{sec:related}


\subsection{Audio-driven talking face generation}
Audio-driven talking face generation, otherwise known as lip-syncing, aims to produce a realistic visual output of a talking head that is accurately synced with the input audio. Neural Voice Puppetry \cite{thies2020neural} is one of the first high-fidelity deepfake pipelines and uses a pretrained speech feature extractor, a per-frame blendshape generator and a neural renderer. FaceFormer \cite{fan2022faceformer} generates a sequence of 3D meshes that match an audio input. CodeTalker \cite{xing2023codetalker} is a similar model that achieves the current state-of-the-art in audio-driven 3D mesh generation.

Models like \cite{paraperas2022ned, eamm} can manipulate videos conditioned on either an emotional label or a reference video. EmoTalk \cite{peng2023emotalk} explicitly disentangles emotional characteristics from audio content and predicts blendshape sequences with expressive emotion. While emotional synthesis is an active research area, most authors note that audio-driven emotional talking head generation lacks control of the input audio, which is naturally correlated with the emotion. 

Works like \cite{stypulkowski2023diffused, shen2023difftalk, bigioi2023speech} leverage the recent advances in diffusion generative models and apply them for high-fidelity talking heads. Explicit style control is another active area of research. TalkCLIP \cite{ma2023talkclip} generates talking faces conditioned either on natural language style prompts or a reference video for guiding the audio generation, thus alleviating the one-to-many mapping and generating diverse samples.

\subsection{Text-driven talking face generation}
Text-driven audiovisual speech synthesis involves the joint generation of audio and visual outputs for a talking face, based on a text input. In this case, the one-to-many mapping is even more challenging to model than in audio-driven approaches. Write-a-speaker \cite{li2021write} generates photorealistic talking head videos from text, with realistic facial expressions and head motions in accordance with speech. Similarly, the authors of \cite{liu2022parallel} propose a parallel model for fast generation of cropped lip image sequences from an input text. However, neither of these models are bimodal, since they only generate visual speech.

Models such as \cite{kumar2017obamanet, wang2022anyonenet, song2022talking, zhang2022text2video, ye2023ada}, use a 2-stage pipeline consisting of a TTS system and an audio-driven talking face module. However, such approaches do not consider the inherent strong correlations of audio and visual streams in human speech. 
The method of \cite{yao2021iterative} also uses text as input, but relies on iterative feedback by the users, in terms of speech wording, mouth motions and performance style.  

DurIAN \cite{yu2019durian} and AVTacotron2 \cite{hussen2021audiovisual} are models with joint audiovisual features that produce talking heads from text. DurIANadapts the WaveRNN model \cite{kalchbrenner2018efficient} to generate animation parameters, while AVTacotron2 expands the Tacotron2 framework \cite{shen2018natural} to produce emotion-conditioned 3DMM sequences. Their major drawback is their autoregressive architecture, which severely impacts the sampling speed and quality of longer samples. Finally, UniFLG \cite{mitsui2023uniflg} learns a joint representation of text and audio, enabling both text-driven and audio-driven synthesis. However, it models the human face with 2D landmarks, which is not a suitable representation for high-detail lip articulation and cannot generalize to new faces. 

In contrast to previous approaches, we introduce a  text-driven audiovisual speech synthesizer that uses transformers and does not follow a cascaded 2-stage approach. Our approach combines advances in TTS and facial reconstruction. 
We rely on the transformer-based FastSpeech 2 \cite{ren2020fastspeech} architecture and use the FLAME 3DMM \cite{li2017learning} for facial modeling.  
By using a joint audiovisual representation, learned with supervision from both modalities, we force the intermediate features to encode better the inherent bimodality of speech. Thus, the representation is closer to how a human would comprehend spoken language. 
We estimate the FLAME parameters using SPECTRE \cite{filntisis2023spectre}, a speech-informed facial reconstruction method that uses a lip-reading loss and is robust to in-the-wild datasets. Thus, our model learns to produce very realistic lip movements in talking head videos. 

\subsection{Photo-realistic facial video synthesis}

Several recent methods synthesize photo-realistic facial videos 
using conditional generative models. 
Methods like \cite{DVP,head2head++,doukas2021headgan} use conditional GANs to render the target subject under the given conditions (expressions, pose, eye-gaze). However, these methods need a driving video of an actor's face and do not offer any semantic control over the generated video.  
This is partly overcome by methods that offer control in terms of facial expressions \cite{Tripathy_ICface,Tripathy_FACEGAN,groth2020altering,DSM21}, without however having any control or constraints on the speech-related facial motions. 
Kim \etal~\cite{neural_style_preserving_dubbing} presented a style-preserving solution to film dubbing, where the expression parameters of the dubber pass through a \textit{style-translation} network before driving the performance of the foreign actor. Their method preserves the dubber's speech, but can only translate between a pair of speaking-styles (dubber-to-actor). 
Papantoniou et al.~\cite{paraperas2022ned} alter the facial expressions of a facial video in a photo-realistic manner while preserving the speech-related lip movements.

In contrast to the aforementioned methods, NEUTART offers the ability to photo-realistically synthesize completely novel lip movements that are in accordance with the input text and simultaneously synthesize novel speech audio that perfectly matches with the synthetic video.

\section{Methodology}
\label{sec:model}

\begin{figure*}[thpb]
  \centering
  \includegraphics[width=\textwidth]{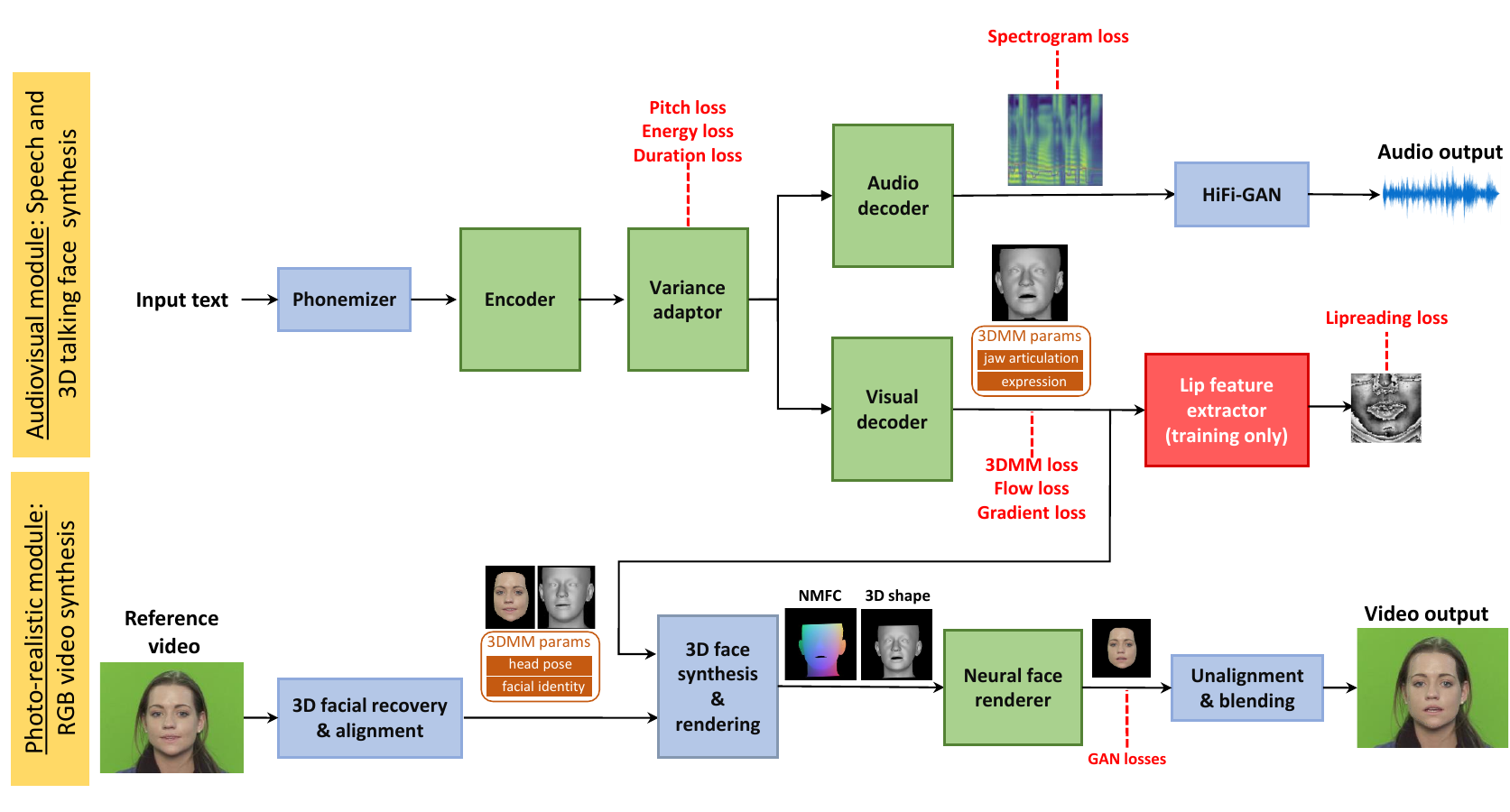}
  \caption{Our model uses two modules for photo-realistic audiovisual speech synthesis. The first module maps the input text to audio as well as a synced 3D talking head. The photo-realistic module swaps the face from a reference video with a face predicted from the 3D talking head. The two modules are coupled during inference, but are trained separately. Green boxes correspond to trainable sub-networks.}
  \label{fig:architecture}
\end{figure*}

\begin{table*}
\centering
\begin{tabular}{@{}llccccccc@{}}
\toprule
ID & Method & MCD (dB) $\downarrow$ & A-CER (\%) $\downarrow$ & LMD $\downarrow$ & LMVE $\downarrow$ & FID $\downarrow$ & V-CER (\%) $\downarrow$ & VER (\%) $\downarrow$ \\
\midrule
\multirow{5}{*}{38F} 
& Ground truth & - & [6.06] & - & - & - & [84.45] & [75.96] \\
& Wav2Lip \cite{prajwal2020lip} & 44.21 & 12.94 & 1.3125 & 0.3238 & \textbf{18.36} & 76.78 & \textbf{68.30} \\
& SadTalker \cite{zhang2022sadtalker} & 44.21 & 12.94 & 14.3464 & 0.4238 & 221.49 & 79.82 & 73.33 \\
& VideoReTalking \cite{cheng2022videoretalking} & 44.21 & 12.94 & 1.6474 & 0.3150 & 37.32 & 80.76 & 75.48 \\
& Ours & \textbf{43.41} & \textbf{10.94} & \textbf{1.1813} & \textbf{0.2889} & 38.14 & \textbf{74.70} & 68.78 \\
\midrule
\multirow{5}{*}{42M} 
& Ground truth & - & [29.64] & - & - & - & [88.68] & [78.52] \\
& Wav2Lip \cite{prajwal2020lip} & 42.58 & \textbf{25.01} & \textbf{1.0609} & \textbf{0.2805} & \textbf{18.45} & 82.21 & 73.81 \\
& SadTalker \cite{zhang2022sadtalker} & 42.58 & \textbf{25.01} & 7.0019 & 0.4010 & 167.47 & 80.42 & 73.09 \\
& VideoReTalking \cite{cheng2022videoretalking} & 42.58 & \textbf{25.01} & 1.5036 & 0.2753 & 25.34 & 78.97 & 71.38 \\
& Ours & \textbf{42.36} & 32.50 & 1.2073 & 0.2867 & 22.91 & \textbf{78.64} & \textbf{71.15} \\
\midrule
\multirow{5}{*}{49F} 
& Ground truth & - & [7.36] & - & - & - & [88.53] & [79.42] \\
& Wav2Lip \cite{prajwal2020lip} & \textbf{43.50} & 18.07 & 1.9305 & 0.4479 & \textbf{17.73} & 87.62 & 81.75 \\
& SadTalker \cite{zhang2022sadtalker} & \textbf{43.50} & 18.07 & 5.3592 & 0.5820 & 139.10 & 83.79 & 77.41 \\
& VideoReTalking \cite{cheng2022videoretalking} & \textbf{43.50} & 18.07 & \textbf{1.9215} & 0.4347 & 29.47 & 84.76 & 78.15 \\
& Ours & 43.64 & \textbf{16.93} & 1.9576 & \textbf{0.4132} & 25.06 & \textbf{76.88} & \textbf{72.02} \\
\midrule
\midrule
\multirow{5}{*}{\textbf{Average}} 
& Ground truth & - & [14.35] & - & - & - & [87.22] & [77.97] \\
& Wav2Lip \cite{prajwal2020lip} & 43.43 & \textbf{18.67} & \textbf{1.4346} & 0.3507 & \textbf{18.18} & 82.20 & 74.62 \\
& SadTalker \cite{zhang2022sadtalker} & 43.43 & \textbf{18.67} & 8.9025 & 0.4689 & 176.02 & 81.34 & 74.61 \\
& VideoReTalking \cite{cheng2022videoretalking} & 43.43 & \textbf{18.67} & 1.6908 & 0.3417 & 30.71 & 81.50 & 75.00 \\
& Ours & \textbf{43.14} & 20.12 & 1.449 & \textbf{0.3296} & 28.70 & \textbf{76.74} & \textbf{70.65} \\
\bottomrule
\end{tabular}
\caption{Generation quality metrics on 3 random, unseen subjects from TCD-TIMIT. The audio evaluation of 42M does not seem comparable to the other subjects, or with the metrics extracted from the multispeaker model, leading us to believe that it is an outlier of the dataset. Nevertheless, our method performs very well in terms of lip landmark metrics and FID. Wav2Lip may have a lower FID, but it performs significantly worse than other methods in terms of human evaluation, due to the visible bounding box in the subject's mouth. Finally, our method is consistently superior when it comes to lipreading.}
\label{tab:generation}
\end{table*}

\begin{table}
\centering
\begin{tabular}{@{}lcc@{}}
\toprule
Method & MCD (dB) $\downarrow$ & ASR-CER (\%) $\downarrow$ \\
\midrule
Ground truth & - & [16.21] \\
FastSpeech 2 \cite{ren2020fastspeech} & \textbf{40.13} & 27.04 \\
Ours & 40.24 & \textbf{24.85} \\
\bottomrule
\end{tabular}
\caption{Multimodality's effect on audio. The audio produced from NEUTART is more intelligible than a plain TTS model of the same architecture, indicating the effectiveness of visual supervision for speech.}
\label{tab:multimodality}
\end{table}

Our Neural Text to Articulate Talk (NEUTART) framework addresses
the challenging problem of 
text-driven, photo-realistic audiovisual speech synthesis. 
An overview of the proposed pipeline at test time is presented in Fig.~\ref{fig:architecture}. It consists of 
two main modules for photo-realistic talking face generation: \textbf{a)} 
an audiovisual module for joint synthesis of speech audio and 3D talking face sequences, and \textbf{b)} a photo-realistic module for synthesis of RGB facial videos. The two modules are coupled during inference, but are trained separately, due to the heavy computational requirements of the neural renderer used in the photo-realistic module. 
Details for each module are presented in the following sections.

\subsection{Audiovisual Module}


Our model builds upon the FastSpeech 2 \cite{ren2020fastspeech} TTS system in order to incorporate visual generation. The audio is modeled via its mel spectrogram, from which a pretrained vocoder produces the speech waveform. Similarly, the faces are modeled with the FLAME 3DMM \cite{li2017learning}, which decouples a face 3D mesh of $5023$ vertices into identity $\boldsymbol{\beta}$, expression $\boldsymbol{\psi}$ and joint pose $\boldsymbol{\theta}$ parameters, which include the 3 jaw articulation parameters $\boldsymbol{\theta}_{jaw}$. Using this representation, we model the facial expression and movements during speech using the $3$ jaw pose and $50$ expression parameters. Thus, NEUTART predicts 80 mel channels and 53 3DMM channels per audio frame. The 3DMM coefficients are then decoded into 3D facial reconstructions, which drive the face renderer. The model consists of the following submodules (see Fig. \ref{fig:architecture}): 
\begin{itemize}
\item \textbf{Phonemizer:} This first submodule takes a plain English text as input and converts it to phoneme sequences according to the CMU pronunciation dictionary. 

\item\textbf{Encoder:} The encoder projects the sequence of phoneme indices into a phoneme embedding sequence, which is processed by a 4-layer transformer. 
\item\textbf{Variance Adaptor:} This submodule is used to add variance to the text encodings. The information modeled consists of three audio features, namely pitch (fundamental frequency), spectrogram energy and phoneme duration. The variance adaptor expands each phoneme encoding according to its duration in mel frames, then adds pitch, energy and liveliness embeddings to the encodings.
\item\textbf{Audio Decoder:} This submodule consists of a 6-layer transformer and a linear layer, which project the intermediate features into the spectrogram.

\item\textbf{Vocoder:} We use the pretrained HiFi-GAN universal generator \cite{kong2020hifi} in order to convert the mel spectrogram into a speech timeseries.

\item\textbf{Visual Decoder:} Similar to the audio decoder, a 4-layer transformer and a linear layer to convert the intermediate features into timeseries of pose and expression coefficients.

\item\textbf{Visual decoder:} A FLAME layer projects the vector $x(t)$ of pose and expression coefficients to a 3D mesh, per video frame $t$. From each 3D mesh, a shape as well a NMFC image are extracted, which are used to drive the renderer.

\end{itemize}

\subsubsection{Training of the Audiovisual Module}

During the training of the audiovisual module, we optimize both decoders' outputs as well as the variance adaptor's predictions. The overall loss function is defined as the simple summation (without any additional external weights to balance the terms) of the following loss terms (see also Fig.~\ref{fig:architecture}):\\\\
\textbf{Pitch, Energy and Duration losses:} These losses correspond to the variance adaptor and are defined as Mean Squared Errors (MSE) of the predicted values, following \cite{ren2020fastspeech}.\\\\
\textbf{Spectrogram loss:} This is related to the audio stream and we define it as the  mean absolute error of the predicted spectrogram, which is the output of the audio decoder.\\\\
\textbf{Visual losses:} The MSE of the predicted FLAME parameters, as well as the blendshape \textit{gradient loss}:
\begin{equation}
\mathcal{L}_{grad} = \frac{1}{T} \sum_t ||x(t+1) - x(t)||_2^2
\end{equation}
and a velocity-type loss that we call \textit{flow loss}:
\begin{equation}
\mathcal{L}_{flow} = \frac{1}{T}  \sum_t ||(\hat x(t+1) - \hat x(t)) - (x(t+1) - x(t))||_2^2
\end{equation}
where $\hat x(t)$ is the actual target value for the FLAME vector and $x(t)$ is the predicted vector at time $t$, for a total of $T$ frames.\\\\
\textbf{Lipreading loss:} Following \cite{filntisis2023spectre}, we use the intermediate features from a pretrained lipreading model \cite{ma2022visual} in order to capture the patterns of lip movements while speaking. Feature vectors are extracted from the ground truth video and the textured 3D mesh, and we use their cosine distance as
\begin{equation}
\mathcal{L}_{lip} = \frac{1}{T} \sum_t \left( 1 - \frac{\vec f_g(t) \cdot \vec f_p(t)}{||\vec f_g(t)||~||\vec f_p(t)||} \right)
\end{equation}
where $\vec f_g(t)$ and $\vec f_p(t)$ are the feature vectors at frame $t$ for the ground truth and predicted video, respectively.\\\\
\textbf{Expression regularization loss:} The Lipreading loss needs an additional constraint on the magnitude of the expression coefficients, otherwise they quickly start oscillating. Thus, along with $\mathcal{L}_{lip}$, we use the following $L2$ regularization loss:
\begin{equation}
\mathcal{L}_{reg} = 10^{-3} \,\, ||\boldsymbol{\psi}(t)||_2
\end{equation}

\subsection{Photo-realistic Module}

Having inferred the corresponding timeseries of jaw and expression parameters from an input transcription, we use them in order to condition a Neural Face Renderer, following  \cite{paraperas2022ned, doukas2021head2head++}. This approach allows us to modify the visual content of an input video featuring a speaking person.




In more detail, the neural renderer is implemented with a convolutional architecture that tackles an image-to-image translation task based on GANs. The input to the neural renderer consists of the concatenation of a rendered 3D face geometry \textbf{S} of the inferred 3D model, with an \textbf{NMFC} (Normalized Mean Face Coordinate) \cite{paraperas2022ned} rendering of the 3D model. At each timestep, we feed the renderer with the current concatenated inputs \textbf{NMFC} and \textbf{S} as well as the previous two frames. 

Note that in contrast to \cite{paraperas2022ned}, we have employed the SPECTRE method for 3D reconstruction \cite{filntisis2023spectre}, which focuses on visual speech-preserving 3D reconstruction. The same 3D reconstructions are used as ground truth values for training the visual decoder in our audiovisual module, ensuring consistency between them.

\subsubsection{Training of the Photo-realistic Module}

The renderer is trained to reconstruct the masked face from the original RGB frame, conditioned on the shape and NMFC images. It employs a GAN-based adversarial loss \cite{goodfellow2014generative}, as well as a specialized mouth discriminator for improved realism in the lip area. 

\section{Experiments}
\label{sec:experiments}

\begin{figure*}[thpb]
\centering
\rotatebox{90}{~~~~~~~ Ours ~~~ VideoReTalking ~ SadTalker ~~ Wav2Lip}
\begin{subfigure}{.48\textwidth}
  \centering
  \includegraphics[width=\textwidth]{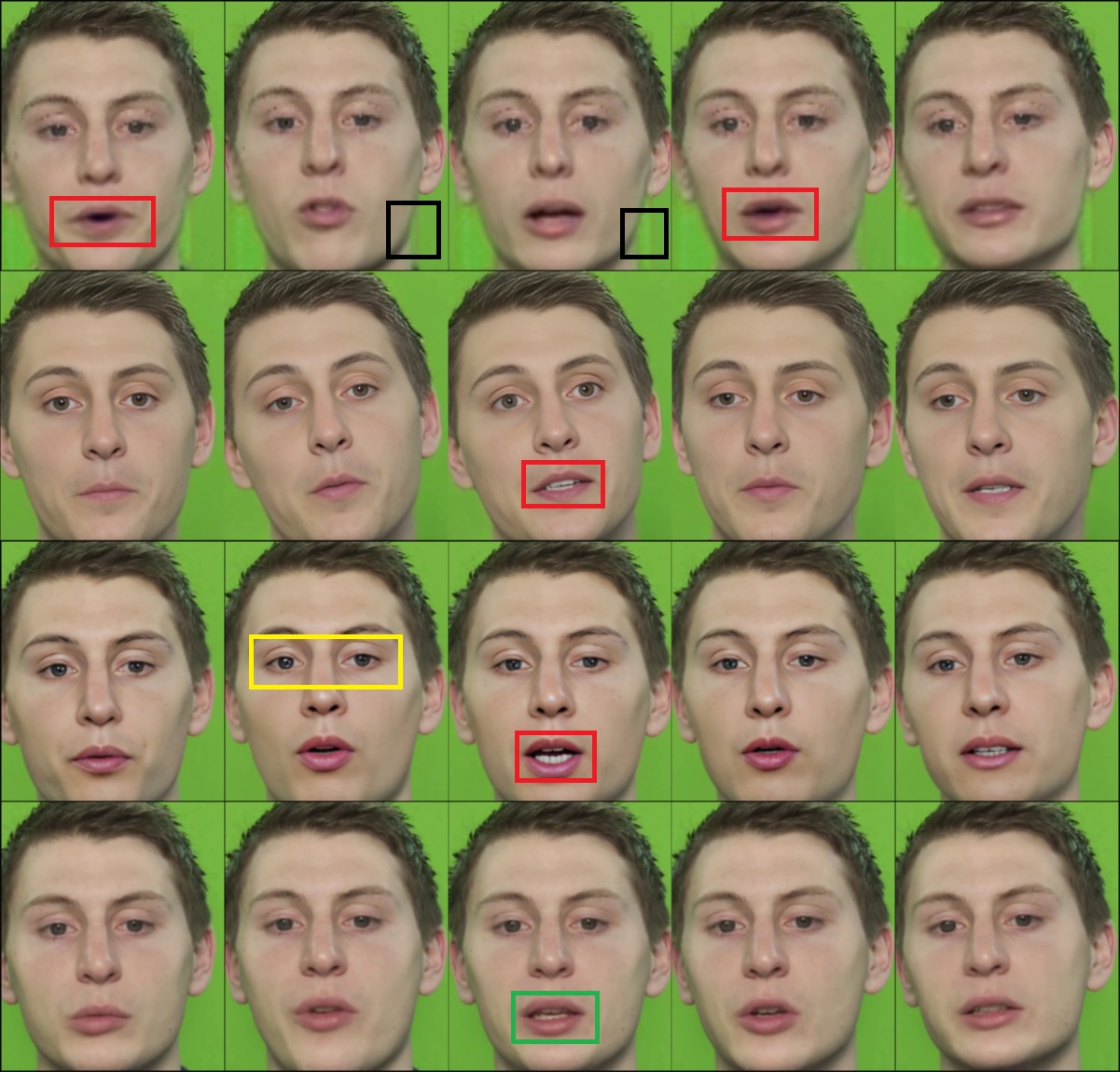}
\end{subfigure}%
~
\begin{subfigure}{.475\textwidth}
  \centering
  \includegraphics[width=\textwidth]{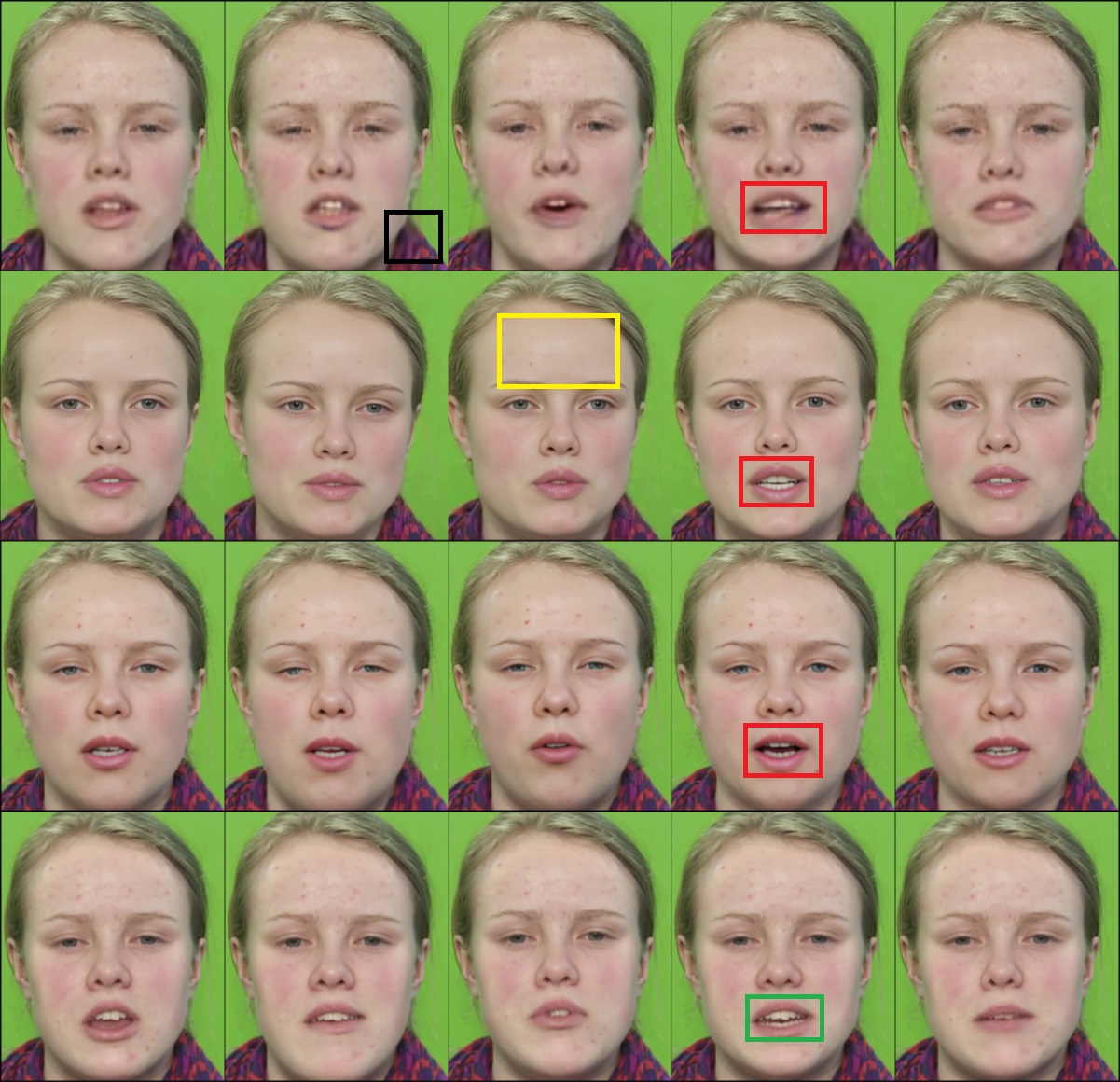}
\end{subfigure}
\caption{Comparison of our method against previous ones. By zooming in, one can see the shortcomings of each method. Wav2Lip has poor resolution, and shows a bounding box artifact around the mouth (black boxes). SadTalker and VideoReTalking produce frames at a much better resolution, since they use a face enhancing network, however this enhancement produces artifacts that alter the person's identity. For instance, the female subject's skin is oversmoothed, while the male subject's eye color is changed (yellow boxes). All three of them produce some frames with unrealistic mouth interior, and seem to alter the subject's teeth (red boxes). Our method generates frames with well-formed lips, teeth and mouth interior, without creating any uncanny effects in the subject's face.}
\label{fig:comparison}
\end{figure*}

We conduct qualitative and quantitative evaluations of our method and comparisons with recent state-of-the-art methods. Additional results and visualizations are provided in the video of the Supp. Material in \cite{project_page}. 

We experimented with the large-scale, multi-speaker audiovisual dataset TCD-TIMIT \cite{harte2015tcd}. Due to the TTS submodule's sensitivity to noise, we initialized the encoder and audio decoder with pretrained weights from the LJSpeech dataset. The visual decoder was randomly initialized, and all the models were trained for 50,000 iterations in each experiment. The TTS alignment is performed by an external aligner before the training phase. We used the Montreal Forced Aligner \cite{mcauliffe2017montreal}. We first trained a multi-speaker model with the TCD-TIMIT data, which can be used as is, or fine-tuned to a particular identity for a few epochs, in order to create person-specific models. The renderer is inherently person-specific, trained on RGB videos of each identity.

\begin{table*}
\centering
\begin{tabular}{@{}ccccc@{}}
\toprule
     &  \multirow{2}{*}{SadTalker}    &  \multirow{2}{*}{VideoRetalking}   & \multirow{2}{*}{Wav2Lip}  & FastSpeech 2\\
     &    &   &  & (audio only)\\
\midrule
\multirow{2}{*}{NEUTART} & \textbf{66}~/~54 & \textbf{74}~/~46 & \textbf{87}~/~33 & \textbf{59}~/~37 \\ 
 & \textbf{55.0\%}~/~45.0\% & \textbf{61.7\%}~/~38.3\% & \textbf{72.5\%}~/~27.5\%  & \textbf{61.5\%}~/~38.5\% \\ 
\bottomrule
\end{tabular}
\caption{
User study preference results: ``A~/~B" indicates NEUTART (left) was preferred \textit{A} times, while the competing method was preferred \textit{B} times (\textit{A}+\textit{B} pairs assessed). 
}
\label{tab:user_study}
\end{table*}

\begin{table*}[t]
\centering
\begin{tabular}{@{}cccccccccc@{}}
\toprule
$\mathcal{L}_{lip}$ & $\mathcal{L}_{grad}$ & $\mathcal{L}_{flow}$ & MCD (dB) $\downarrow$ & A-CER (\%) $\downarrow$ & LMD $\downarrow$ & LMVE $\downarrow$ & V-CER (\%) $\downarrow$ & VER (\%) $\downarrow$ \\
\midrule
\xmark & \xmark & \cmark & 41.98 & \textbf{21.91} & 0.5053 & 0.3502 & 82.40 & 77.90 \\
\xmark & \cmark & \xmark & 41.91 & 22.88 & 0.6856 & 0.3602 & 85.66 & 80.00 \\
\xmark & \cmark & \cmark & \textbf{40.31} & 25.05 & \textbf{0.4318} & \textbf{0.3261} & 77.15 & 70.99 \\
\cmark & \cmark & \cmark & \textbf{40.31} & 25.40 & 0.5063 & 0.4203 & \textbf{77.05} & \textbf{70.66} \\
\bottomrule
\end{tabular}
\caption{Ablation study on the effectiveness of visual losses. Apart from the lipreading loss which increases the articulation realism, the usage of gradient and flow losses ensures more accurate landmark prediction. As a result, we include them for training the audiovisual module.}
\label{tab:ablation}
\end{table*}

\subsection{Evaluation}
Evaluation for talking faces is a challenge, since statistical error measures do not correlate very well with human assessments \cite{chen2020comprises}. We objectively evaluate our model using both statistical and perceptual metrics across different modalities, and also conduct a user study. 

\subsubsection{Objective evaluation}
The objective evaluation is performed on 3 randomly chosen subjects from the TCD-TIMIT dataset, unseen during the training of the multispeaker model. We use person-specific audiovisual and renderer modules and compare NEUTART's results with three open-sourced methods, in order to obtain reproducible results. Since there are not any recent publicly available text-driven methods, we choose to compare our model against the following popular lip-syncing models: Wav2Lip \cite{prajwal2020lip}, SadTalker \cite{zhang2022sadtalker} and VideoReTalking \cite{cheng2022videoretalking}. The latter achieves state-of-the-art results in audio-driven photorealistic talking head generation. Since we are targeting audiovisual generation from text, we sampled the audio-driven models using audios from FastSpeech 2, which is equivalent to our TTS submodule. Since the other methods are either one-shot or few-shot, we used real videos from TCD-TIMIT as reference for the generation.

For evaluating the audio, we use the average mel cepstral distance (MCD), which for a predicted spectrogram $S$ is:
\begin{equation}
MCD = 10 \log_{10} ||\hat S - S||_2^2
\end{equation}
We also perceptually evaluate the audio quality using the character error rate (A-CER) from the Wav2Vec2 ASR model \cite{baevski2020wav2vec}. The visual articulation quality is assessed by the average lip landmark distance (LMD) and the average lip landmark velocity error (LLVE) between the predicted video and the original. The landmarks are extracted with Google's MediaPipe. Following ~\cite{filntisis2023spectre}, we perform visual perceptual evaluation by cropping the mouth videos and using the pretrained AV-HuBERT \cite{shi2022avhubert, shi2022avsr} as a lip-reader. The metrics we use are the visual character error rate and the and viseme error rate, after mapping the phonemes into visemes.  Finally, the photo-realism is evaluated with Fréchet inception distance (FID) \cite{heusel2017gans}. Note that calculating some of the aforementioned errors requires the compared sequences to be of the same length, therefore we align them using dynamic time warping \cite{muller2007dynamic}. We present the results in Table \ref{tab:generation}.

The results indicate that NEUTART can produce articulate and coherent talking heads. We also compare NEUTART's performance on audio synthesis compared to an equivalent TTS system (FastSpeech 2) and present the metrics in Table \ref{tab:multimodality}. The results indicate that including a visual visual supervision can improve the generated audio's quality, especially in terms of intelligibility, thus supporting the effectiveness of multitask learning.

\subsubsection{Subjective evaluation}
We also conducted a user study, comparing our method against FastSpeech 2 in terms of audio realism, and all the aforementioned audio-driven methods in terms of audiovisual realism. To do this, we first selected a set of unseen phonetically rich text transcriptions to generate the results for each method, using two randomly chosen subjects (one male and one female). We used subjects from TCD-TIMIT and not in-the-wild, since our renderer turns out to be sensitive to big variations in head pose, which are common in natural speech. For the study, we adopted a preference test design. For each audio-based question users were presented with two audio files and were asked to select the one that sounds more realistic. For the audiovisual part, two synthetic videos where presented and the users were asked again to select the one that they find more realistic. Note that in each question we provided users with the transcription of the audio/video. In the case of video we also provided users with an image of the synthetically generated person from the original footage.
Each user answered a total of $4$ audio-based questions and $15$ video-based questions (5 questions for each audiovisual pair - ours versus another method). A total of 21 users completed the questionnaire and the results can be seen in  Table~\ref{tab:user_study}. We see that our method is consistently perceived as more realistic by independent users. SadTalker is evaluated as the second best method, then follows VideoReTalking. We also revisit the effectiveness of multitask learning for TTS, by showing that the audiovisual model generates speech that is preferable to a plain TTS system's output. The first sample from each of the two subjects is presented in Fig.~\ref{fig:comparison}, where we annotate key shortcomings of previous methods.

\subsection{Ablation study}
In order to study the effects of each additional visual loss on the talking face generation, we performed an ablation study on the TCD-TIMIT dataset. The audiovisual generator's goal is to synthesize 3D faces as accurately as possible, so we evaluated the generation in terms of audio and 3D reconstruction, with visual metrics computed from images of 3D reconstructions. We present the results in Table \ref{tab:ablation}. Most metrics are lower when the model is trained using all visual losses, which indicates their effectiveness.

\section{Conclusions}
\label{sec:cxperiments}

\subsection{Summary}
In this paper, we presented NEUTART, a photorealistic text-driven audiovisual speech generator trained on accurate 3D facial reconstructions, as well as a perceptual lip-reading loss for visual supervision. The model is built with a transformer-based encoder-decoder architecture and uses joint audiovisual encodings mapped from a phoneme sequence. Thus, the audiovisual correlation is better captured, while alleviating the redundancy of extracting features from generated speech in order to create a video. Our experiments show that our model can generate samples that achieve both auditory and visual realism, especially in terms of speech articulation.

\subsection{Future Work}
Performance is usually disregarded in talking face generation, especially in research, whereas low computational cost and inference speed are essential for commercial applications and mobile devices. In NEUTART, the slowest component is the neural renderer, which operates in the video's pixel space. In future versions of our work, we plan to optimize the renderer's architecture in order to achieve faster sampling. Moreover, employing end-to-end training by backpropagating the image-space error into the audiovisual submodule can further improve the overall realism of the generated output.

\subsection{Ethical Statement}
As already mentioned, deep learning systems for photorealistic audiovisual speech synthesis like ours can have a very positive impact in many applications such as digital avatars, virtual assistants, accessibility tools, teleconferencing, video games, movie dubbing, and human-machine interfaces. However, at the same time, this type of technology has the risk of being misused towards unethical or malicious purposes, since it can produce deepfake photorealistic videos of individuals (\eg~celebrities or politicians) saying things that they have never said, without their consent. This raises concerns related to the creation and distribution of fake news and other negative social impact \cite{chesney2019deep,diakopoulos2021anticipating,yadlin2021whose}. 
We believe that researchers and engineers working in this field need to be mindful of these ethical issues and contribute to raising public awareness about the capabilities of such AI systems.  
Other countermeasures include contributing in developing state-of-the-art systems that detect deepfake videos \cite{zhang2022deepfake,masood2023deepfakes}.

{
    \small
    \bibliographystyle{ieeenat_fullname}
    \bibliography{main}

\begin{thebibliography}{57}
\providecommand{\natexlab}[1]{#1}
\providecommand{\url}[1]{\texttt{#1}}
\expandafter\ifx\csname urlstyle\endcsname\relax
  \providecommand{\doi}[1]{doi: #1}\else
  \providecommand{\doi}{doi: \begingroup \urlstyle{rm}\Url}\fi

\bibitem[pro()]{project_page}
\url{https://g-milis.github.io/neutart.html}.

\bibitem[Abdelaziz et~al.(2021)Abdelaziz, Kumar, Seivwright, Fanelli, Binder, Stylianou, and Kajareker]{hussen2021audiovisual}
Ahmed~Hussen Abdelaziz, Anushree~Prasanna Kumar, Chloe Seivwright, Gabriele Fanelli, Justin Binder, Yannis Stylianou, and Sachin Kajareker.
\newblock Audiovisual speech synthesis using tacotron2.
\newblock In \emph{Proceedings of the 2021 International Conference on Multimodal Interaction}, pages 503--511, 2021.

\bibitem[Baevski et~al.(2020)Baevski, Zhou, Mohamed, and Auli]{baevski2020wav2vec}
Alexei Baevski, Yuhao Zhou, Abdelrahman Mohamed, and Michael Auli.
\newblock wav2vec 2.0: A framework for self-supervised learning of speech representations.
\newblock \emph{Advances in neural information processing systems}, 33:\penalty0 12449--12460, 2020.

\bibitem[Bigioi et~al.(2023)Bigioi, Basak, Jordan, McDonnell, and Corcoran]{bigioi2023speech}
Dan Bigioi, Shubhajit Basak, Hugh Jordan, Rachel McDonnell, and Peter Corcoran.
\newblock Speech driven video editing via an audio-conditioned diffusion model.
\newblock \emph{arXiv preprint arXiv:2301.04474}, 2023.

\bibitem[Chen et~al.(2020)Chen, Cui, Kou, Zheng, and Xu]{chen2020comprises}
Lele Chen, Guofeng Cui, Ziyi Kou, Haitian Zheng, and Chenliang Xu.
\newblock What comprises a good talking-head video generation?
\newblock In \emph{IEEE/CVF Conference on Computer Vision and Pattern Recognition Workshops}, 2020.

\bibitem[Cheng et~al.(2022)Cheng, Cun, Zhang, Xia, Yin, Zhu, Wang, Wang, and Wang]{cheng2022videoretalking}
Kun Cheng, Xiaodong Cun, Yong Zhang, Menghan Xia, Fei Yin, Mingrui Zhu, Xuan Wang, Jue Wang, and Nannan Wang.
\newblock Videoretalking: Audio-based lip synchronization for talking head video editing in the wild.
\newblock In \emph{SIGGRAPH Asia 2022 Conference Papers}, pages 1--9, 2022.

\bibitem[Chesney and Citron(2019)]{chesney2019deep}
Bobby Chesney and Danielle Citron.
\newblock Deep fakes: A looming challenge for privacy, democracy, and national security.
\newblock \emph{Calif. L. Rev.}, 107:\penalty0 1753, 2019.

\bibitem[Diakopoulos and Johnson(2021)]{diakopoulos2021anticipating}
Nicholas Diakopoulos and Deborah Johnson.
\newblock Anticipating and addressing the ethical implications of deepfakes in the context of elections.
\newblock \emph{New Media \& Society}, 23\penalty0 (7):\penalty0 2072--2098, 2021.

\bibitem[Doukas et~al.(2021)Doukas, Koujan, Sharmanska, Roussos, and Zafeiriou]{doukas2021head2head++}
Michail~Christos Doukas, Mohammad~Rami Koujan, Viktoriia Sharmanska, Anastasios Roussos, and Stefanos Zafeiriou.
\newblock Head2head++: Deep facial attributes re-targeting.
\newblock \emph{IEEE Transactions on Biometrics, Behavior, and Identity Science}, 3\penalty0 (1):\penalty0 31--43, 2021.

\bibitem[{Doukas} et~al.(2021){Doukas}, {Koujan}, {Sharmanska}, {Roussos}, and {Zafeiriou}]{head2head++}
Michail~Christos {Doukas}, Mohammad~Rami {Koujan}, Viktoriia {Sharmanska}, Anastasios {Roussos}, and Stefanos {Zafeiriou}.
\newblock Head2head++: Deep facial attributes re-targeting.
\newblock \emph{IEEE Transactions on Biometrics, Behavior, and Identity Science}, 3\penalty0 (1):\penalty0 31--43, 2021.

\bibitem[Doukas et~al.(2021)Doukas, Zafeiriou, and Sharmanska]{doukas2021headgan}
Michail~Christos Doukas, Stefanos Zafeiriou, and Viktoriia Sharmanska.
\newblock Headgan: One-shot neural head synthesis and editing.
\newblock In \emph{Proceedings of the IEEE/CVF International conference on Computer Vision}, pages 14398--14407, 2021.

\bibitem[Fan et~al.(2022)Fan, Lin, Saito, Wang, and Komura]{fan2022faceformer}
Yingruo Fan, Zhaojiang Lin, Jun Saito, Wenping Wang, and Taku Komura.
\newblock Faceformer: Speech-driven 3d facial animation with transformers.
\newblock In \emph{Proceedings of the IEEE/CVF Conference on Computer Vision and Pattern Recognition}, pages 18770--18780, 2022.

\bibitem[Filntisis et~al.(2023)Filntisis, Retsinas, Paraperas-Papantoniou, Katsamanis, Roussos, and Maragos]{filntisis2023spectre}
Panagiotis~P Filntisis, George Retsinas, Foivos Paraperas-Papantoniou, Athanasios Katsamanis, Anastasios Roussos, and Petros Maragos.
\newblock Spectre: Visual speech-informed perceptual 3d facial expression reconstruction from videos.
\newblock In \emph{Proceedings of the IEEE/CVF Conference on Computer Vision and Pattern Recognition}, pages 5744--5754, 2023.

\bibitem[Goodfellow et~al.(2014)Goodfellow, Pouget-Abadie, Mirza, Xu, Warde-Farley, Ozair, Courville, and Bengio]{goodfellow2014generative}
Ian Goodfellow, Jean Pouget-Abadie, Mehdi Mirza, Bing Xu, David Warde-Farley, Sherjil Ozair, Aaron Courville, and Yoshua Bengio.
\newblock Generative adversarial nets.
\newblock In \emph{Advances in Neural Information Processing Systems}. Curran Associates, Inc., 2014.

\bibitem[Groth et~al.(2020)Groth, Tauscher, Castillo, and Magnor]{groth2020altering}
Colin Groth, Jan-Philipp Tauscher, Susana Castillo, and Marcus Magnor.
\newblock Altering the conveyed facial emotion through automatic reenactment of video portraits.
\newblock In \emph{Proceedings of the International Conference on Computer Animation and Social Agents ({CASA})}, pages 128--135, 2020.

\bibitem[Harte and Gillen(2015)]{harte2015tcd}
Naomi Harte and Eoin Gillen.
\newblock Tcd-timit: An audio-visual corpus of continuous speech.
\newblock \emph{IEEE Transactions on Multimedia}, 17\penalty0 (5):\penalty0 603--615, 2015.

\bibitem[Heusel et~al.(2017)Heusel, Ramsauer, Unterthiner, Nessler, and Hochreiter]{heusel2017gans}
Martin Heusel, Hubert Ramsauer, Thomas Unterthiner, Bernhard Nessler, and Sepp Hochreiter.
\newblock Gans trained by a two time-scale update rule converge to a local nash equilibrium.
\newblock \emph{Advances in neural information processing systems}, 30, 2017.

\bibitem[Ji et~al.(2022)Ji, Zhou, Wang, Wu, Wu, Xu, and Cao]{eamm}
Xinya Ji, Hang Zhou, Kaisiyuan Wang, Qianyi Wu, Wayne Wu, Feng Xu, and Xun Cao.
\newblock Eamm: One-shot emotional talking face via audio-based emotion-aware motion model.
\newblock In \emph{ACM SIGGRAPH 2022 Conference Proceedings}, 2022.

\bibitem[Kalchbrenner et~al.(2018)Kalchbrenner, Elsen, Simonyan, Noury, Casagrande, Lockhart, Stimberg, Oord, Dieleman, and Kavukcuoglu]{kalchbrenner2018efficient}
Nal Kalchbrenner, Erich Elsen, Karen Simonyan, Seb Noury, Norman Casagrande, Edward Lockhart, Florian Stimberg, Aaron Oord, Sander Dieleman, and Koray Kavukcuoglu.
\newblock Efficient neural audio synthesis.
\newblock In \emph{International Conference on Machine Learning}, pages 2410--2419. PMLR, 2018.

\bibitem[Kim et~al.(2018)Kim, Garrido, Tewari, Xu, Thies, Nie{\ss}ner, P{\'e}rez, Richardt, Zoll{\"o}fer, and Theobalt]{DVP}
Hyeongwoo Kim, Pablo Garrido, Ayush Tewari, Weipeng Xu, Justus Thies, Matthias Nie{\ss}ner, Patrick P{\'e}rez, Christian Richardt, Michael Zoll{\"o}fer, and Christian Theobalt.
\newblock Deep video portraits.
\newblock \emph{ACM Transactions on Graphics (TOG)}, 37\penalty0 (4):\penalty0 163, 2018.

\bibitem[Kim et~al.(2019)Kim, Elgharib, Zollh\"{o}fer, Seidel, Beeler, Richardt, and Theobalt]{neural_style_preserving_dubbing}
Hyeongwoo Kim, Mohamed Elgharib, Michael Zollh\"{o}fer, Hans-Peter Seidel, Thabo Beeler, Christian Richardt, and Christian Theobalt.
\newblock Neural style-preserving visual dubbing.
\newblock \emph{ACM Trans. Graph.}, 38\penalty0 (6), 2019.

\bibitem[Kong et~al.(2020)Kong, Kim, and Bae]{kong2020hifi}
Jungil Kong, Jaehyeon Kim, and Jaekyoung Bae.
\newblock Hifi-gan: Generative adversarial networks for efficient and high fidelity speech synthesis.
\newblock \emph{Advances in Neural Information Processing Systems}, 33:\penalty0 17022--17033, 2020.

\bibitem[Kumar et~al.(2017)Kumar, Sotelo, Kumar, de~Br{\'e}bisson, and Bengio]{kumar2017obamanet}
Rithesh Kumar, Jose Sotelo, Kundan Kumar, Alexandre de Br{\'e}bisson, and Yoshua Bengio.
\newblock Obamanet: Photo-realistic lip-sync from text.
\newblock \emph{arXiv preprint arXiv:1801.01442}, 2017.

\bibitem[Li et~al.(2021)Li, Wang, Zhang, Ding, Zheng, Yu, and Fan]{li2021write}
Lincheng Li, Suzhen Wang, Zhimeng Zhang, Yu Ding, Yixing Zheng, Xin Yu, and Changjie Fan.
\newblock Write-a-speaker: Text-based emotional and rhythmic talking-head generation.
\newblock In \emph{Proceedings of the AAAI Conference on Artificial Intelligence}, pages 1911--1920, 2021.

\bibitem[Li et~al.(2017)Li, Bolkart, Black, Li, and Romero]{li2017learning}
Tianye Li, Timo Bolkart, Michael~J Black, Hao Li, and Javier Romero.
\newblock Learning a model of facial shape and expression from 4d scans.
\newblock \emph{ACM Trans. Graph.}, 36\penalty0 (6):\penalty0 194--1, 2017.

\bibitem[Liu et~al.(2022)Liu, Zhu, Ren, Huang, Huai, Yuan, and Zhao]{liu2022parallel}
Jinglin Liu, Zhiying Zhu, Yi Ren, Wencan Huang, Baoxing Huai, Nicholas Yuan, and Zhou Zhao.
\newblock Parallel and high-fidelity text-to-lip generation.
\newblock In \emph{Proceedings of the AAAI Conference on Artificial Intelligence}, pages 1738--1746, 2022.

\bibitem[Ma et~al.(2022)Ma, Petridis, and Pantic]{ma2022visual}
Pingchuan Ma, Stavros Petridis, and Maja Pantic.
\newblock {Visual Speech Recognition for Multiple Languages in the Wild}.
\newblock \emph{{Nature Machine Intelligence}}, 4:\penalty0 930--939, 2022.

\bibitem[Ma et~al.(2023)Ma, Wang, Ding, Ma, Lv, Fan, Hu, Deng, and Yu]{ma2023talkclip}
Yifeng Ma, Suzhen Wang, Yu Ding, Bowen Ma, Tangjie Lv, Changjie Fan, Zhipeng Hu, Zhidong Deng, and Xin Yu.
\newblock Talkclip: Talking head generation with text-guided expressive speaking styles.
\newblock \emph{arXiv preprint arXiv:2304.00334}, 2023.

\bibitem[Masood et~al.(2023)Masood, Nawaz, Malik, Javed, Irtaza, and Malik]{masood2023deepfakes}
Momina Masood, Mariam Nawaz, Khalid~Mahmood Malik, Ali Javed, Aun Irtaza, and Hafiz Malik.
\newblock Deepfakes generation and detection: State-of-the-art, open challenges, countermeasures, and way forward.
\newblock \emph{Applied intelligence}, 53\penalty0 (4):\penalty0 3974--4026, 2023.

\bibitem[McAuliffe et~al.(2017)McAuliffe, Socolof, Mihuc, Wagner, and Sonderegger]{mcauliffe2017montreal}
Michael McAuliffe, Michaela Socolof, Sarah Mihuc, Michael Wagner, and Morgan Sonderegger.
\newblock Montreal forced aligner: Trainable text-speech alignment using kaldi.
\newblock In \emph{Interspeech}, pages 498--502, 2017.

\bibitem[Mitsui et~al.(2023)Mitsui, Hono, and Sawada]{mitsui2023uniflg}
Kentaro Mitsui, Yukiya Hono, and Kei Sawada.
\newblock Uniflg: Unified facial landmark generator from text or speech.
\newblock \emph{arXiv preprint arXiv:2302.14337}, 2023.

\bibitem[M{\"u}ller(2007)]{muller2007dynamic}
Meinard M{\"u}ller.
\newblock Dynamic time warping.
\newblock \emph{Information retrieval for music and motion}, pages 69--84, 2007.

\bibitem[Paraperas~Papantoniou et~al.(2022)Paraperas~Papantoniou, Filntisis, Maragos, and Roussos]{paraperas2022ned}
Foivos Paraperas~Papantoniou, Panagiotis~P. Filntisis, Petros Maragos, and Anastasios Roussos.
\newblock Neural emotion director: Speech-preserving semantic control of facial expressions in "in-the-wild" videos.
\newblock In \emph{Proceedings of the IEEE/CVF Conference on Computer Vision and Pattern Recognition (CVPR)}, 2022.

\bibitem[Peng et~al.(2023)Peng, Wu, Song, Xu, Zhu, Liu, He, and Fan]{peng2023emotalk}
Ziqiao Peng, Haoyu Wu, Zhenbo Song, Hao Xu, Xiangyu Zhu, Hongyan Liu, Jun He, and Zhaoxin Fan.
\newblock Emotalk: Speech-driven emotional disentanglement for 3d face animation.
\newblock \emph{arXiv preprint arXiv:2303.11089}, 2023.

\bibitem[Prajwal et~al.(2020)Prajwal, Mukhopadhyay, Namboodiri, and Jawahar]{prajwal2020lip}
KR Prajwal, Rudrabha Mukhopadhyay, Vinay~P Namboodiri, and CV Jawahar.
\newblock A lip sync expert is all you need for speech to lip generation in the wild.
\newblock In \emph{Proceedings of the 28th ACM International Conference on Multimedia}, pages 484--492, 2020.

\bibitem[Ren et~al.(2020)Ren, Hu, Tan, Qin, Zhao, Zhao, and Liu]{ren2020fastspeech}
Yi Ren, Chenxu Hu, Xu Tan, Tao Qin, Sheng Zhao, Zhou Zhao, and Tie-Yan Liu.
\newblock Fastspeech 2: Fast and high-quality end-to-end text to speech.
\newblock In \emph{International Conference on Learning Representations}, 2020.

\bibitem[Shen et~al.(2018)Shen, Pang, Weiss, Schuster, Jaitly, Yang, Chen, Zhang, Wang, Skerrv-Ryan, et~al.]{shen2018natural}
Jonathan Shen, Ruoming Pang, Ron~J Weiss, Mike Schuster, Navdeep Jaitly, Zongheng Yang, Zhifeng Chen, Yu Zhang, Yuxuan Wang, Rj Skerrv-Ryan, et~al.
\newblock Natural tts synthesis by conditioning wavenet on mel spectrogram predictions.
\newblock In \emph{2018 IEEE international conference on acoustics, speech and signal processing (ICASSP)}, pages 4779--4783. IEEE, 2018.

\bibitem[Shen et~al.(2023)Shen, Zhao, Meng, Li, Zhu, Zhou, and Lu]{shen2023difftalk}
Shuai Shen, Wenliang Zhao, Zibin Meng, Wanhua Li, Zheng Zhu, Jie Zhou, and Jiwen Lu.
\newblock Difftalk: Crafting diffusion models for generalized talking head synthesis.
\newblock In \emph{arxiv}, 2023.

\bibitem[Shi et~al.(2022{\natexlab{a}})Shi, Hsu, Lakhotia, and Mohamed]{shi2022avhubert}
Bowen Shi, Wei-Ning Hsu, Kushal Lakhotia, and Abdelrahman Mohamed.
\newblock Learning audio-visual speech representation by masked multimodal cluster prediction.
\newblock \emph{arXiv preprint arXiv:2201.02184}, 2022{\natexlab{a}}.

\bibitem[Shi et~al.(2022{\natexlab{b}})Shi, Hsu, and Mohamed]{shi2022avsr}
Bowen Shi, Wei-Ning Hsu, and Abdelrahman Mohamed.
\newblock Robust self-supervised audio-visual speech recognition.
\newblock \emph{arXiv preprint arXiv:2201.01763}, 2022{\natexlab{b}}.

\bibitem[Solanki and Roussos(2021)]{DSM21}
Girish~Kumar Solanki and Anastasios Roussos.
\newblock Deep semantic manipulation of facial videos.
\newblock \emph{arXiv preprint arXiv:2111.07902}, 2021.

\bibitem[Song et~al.(2022)Song, Woo, Lee, Yang, Cho, Lee, Choi, and Kim]{song2022talking}
Hyoung-Kyu Song, Sang~Hoon Woo, Junhyeok Lee, Seungmin Yang, Hyunjae Cho, Youseong Lee, Dongho Choi, and Kang-wook Kim.
\newblock Talking face generation with multilingual tts.
\newblock In \emph{Proceedings of the IEEE/CVF Conference on Computer Vision and Pattern Recognition}, pages 21425--21430, 2022.

\bibitem[Stypu{\l}kowski et~al.(2023)Stypu{\l}kowski, Vougioukas, He, Zieba, Petridis, and Pantic]{stypulkowski2023diffused}
Micha{\l} Stypu{\l}kowski, Konstantinos Vougioukas, Sen He, Maciej Zieba, Stavros Petridis, and Maja Pantic.
\newblock Diffused heads: Diffusion models beat gans on talking-face generation.
\newblock \emph{arXiv preprint arXiv:2301.03396}, 2023.

\bibitem[Tan et~al.(2021)Tan, Qin, Soong, and Liu]{tan2021survey}
Xu Tan, Tao Qin, Frank Soong, and Tie-Yan Liu.
\newblock A survey on neural speech synthesis.
\newblock \emph{arXiv preprint arXiv:2106.15561}, 2021.

\bibitem[Thies et~al.(2020)Thies, Elgharib, Tewari, Theobalt, and Nie{\ss}ner]{thies2020neural}
Justus Thies, Mohamed Elgharib, Ayush Tewari, Christian Theobalt, and Matthias Nie{\ss}ner.
\newblock Neural voice puppetry: Audio-driven facial reenactment.
\newblock In \emph{Computer Vision--ECCV 2020: 16th European Conference, Glasgow, UK, August 23--28, 2020, Proceedings, Part XVI 16}, pages 716--731. Springer, 2020.

\bibitem[Toshpulatov et~al.(2023)Toshpulatov, Lee, and Lee]{toshpulatov2023talking}
Mukhiddin Toshpulatov, Wookey Lee, and Suan Lee.
\newblock Talking human face generation: A survey.
\newblock \emph{Expert Systems with Applications}, page 119678, 2023.

\bibitem[Tripathy et~al.(2020)Tripathy, Kannala, and Rahtu]{Tripathy_ICface}
Soumya Tripathy, Juho Kannala, and Esa Rahtu.
\newblock Icface: Interpretable and controllable face reenactment using gans.
\newblock In \emph{Proceedings of the IEEE/CVF Winter Conference on Applications of Computer Vision (WACV)}, 2020.

\bibitem[Tripathy et~al.(2021)Tripathy, Kannala, and Rahtu]{Tripathy_FACEGAN}
Soumya Tripathy, Juho Kannala, and Esa Rahtu.
\newblock Facegan: Facial attribute controllable reenactment gan.
\newblock In \emph{Proceedings of the IEEE/CVF Winter Conference on Applications of Computer Vision (WACV)}, 2021.

\bibitem[Wang et~al.(2022)Wang, Xie, Zhu, Xie, and Scharenborg]{wang2022anyonenet}
Xinsheng Wang, Qicong Xie, Jihua Zhu, Lei Xie, and Odette Scharenborg.
\newblock Anyonenet: Synchronized speech and talking head generation for arbitrary persons.
\newblock \emph{IEEE Transactions on Multimedia}, 2022.

\bibitem[Xing et~al.(2023)Xing, Xia, Zhang, Cun, Wang, and Wong]{xing2023codetalker}
Jinbo Xing, Menghan Xia, Yuechen Zhang, Xiaodong Cun, Jue Wang, and Tien-Tsin Wong.
\newblock Codetalker: Speech-driven 3d facial animation with discrete motion prior.
\newblock \emph{arXiv preprint arXiv:2301.02379}, 2023.

\bibitem[Yadlin-Segal and Oppenheim(2021)]{yadlin2021whose}
Aya Yadlin-Segal and Yael Oppenheim.
\newblock Whose dystopia is it anyway? deepfakes and social media regulation.
\newblock \emph{Convergence}, 27\penalty0 (1):\penalty0 36--51, 2021.

\bibitem[Yao et~al.(2021)Yao, Fried, Fatahalian, and Agrawala]{yao2021iterative}
Xinwei Yao, Ohad Fried, Kayvon Fatahalian, and Maneesh Agrawala.
\newblock Iterative text-based editing of talking-heads using neural retargeting.
\newblock \emph{ACM Transactions on Graphics (TOG)}, 40\penalty0 (3):\penalty0 1--14, 2021.

\bibitem[Ye et~al.(2023)Ye, Jiang, Ren, Liu, Zhang, Yin, Ma, and Zhao]{ye2023ada}
Zhenhui Ye, Ziyue Jiang, Yi Ren, Jinglin Liu, Chen Zhang, Xiang Yin, Zejun Ma, and Zhou Zhao.
\newblock Ada-tta: Towards adaptive high-quality text-to-talking avatar synthesis.
\newblock \emph{arXiv preprint arXiv:2306.03504}, 2023.

\bibitem[Yu et~al.(2019)Yu, Lu, Hu, Yu, Weng, Xu, Liu, Tuo, Kang, Lei, et~al.]{yu2019durian}
Chengzhu Yu, Heng Lu, Na Hu, Meng Yu, Chao Weng, Kun Xu, Peng Liu, Deyi Tuo, Shiyin Kang, Guangzhi Lei, et~al.
\newblock Durian: Duration informed attention network for multimodal synthesis.
\newblock \emph{arXiv preprint arXiv:1909.01700}, 2019.

\bibitem[Zhang et~al.(2022{\natexlab{a}})Zhang, Yuan, Liao, and Zhang]{zhang2022text2video}
Sibo Zhang, Jiahong Yuan, Miao Liao, and Liangjun Zhang.
\newblock Text2video: Text-driven talking-head video synthesis with personalized phoneme-pose dictionary.
\newblock In \emph{ICASSP 2022-2022 IEEE International Conference on Acoustics, Speech and Signal Processing (ICASSP)}, pages 2659--2663. IEEE, 2022{\natexlab{a}}.

\bibitem[Zhang(2022)]{zhang2022deepfake}
Tao Zhang.
\newblock Deepfake generation and detection, a survey.
\newblock \emph{Multimedia Tools and Applications}, 81\penalty0 (5):\penalty0 6259--6276, 2022.

\bibitem[Zhang et~al.(2022{\natexlab{b}})Zhang, Cun, Wang, Zhang, Shen, Guo, Shan, and Wang]{zhang2022sadtalker}
Wenxuan Zhang, Xiaodong Cun, Xuan Wang, Yong Zhang, Xi Shen, Yu Guo, Ying Shan, and Fei Wang.
\newblock Sadtalker: Learning realistic 3d motion coefficients for stylized audio-driven single image talking face animation.
\newblock \emph{arXiv preprint arXiv:2211.12194}, 2022{\natexlab{b}}.

\end{thebibliography}
}


\end{document}